# A Multimodal RAG Framework for Housing Damage Assessment: Collaborative Optimization of Image Encoding and Policy Vector Retrieval

Jiayi Miao[*]

University of California, Berkeley, Blum Center for Developing Economies, Berkeley, USA, jiayi.miao@berkeley.edu

Dingxin Lu

Icahn School of Medicine at Mount Sinai, New York, USA, sydneylu1998@gmail.com

Zhuqi Wang

McKelvey School of Engineering, Washington University in St. Louis, St. Louis, USA, zhuqi.wang@outlook.com

**Abstract**

After natural disasters, accurate evaluations of damage to housing are important for insurance claims response and planning of resources. In this work, we introduce a novel multimodal retrieval-augmented generation (MM-RAG) framework. On top of classical RAG architecture, we further the framework to devise a two-branch multimodal encoder structure that the image branch employs a visual encoder composed of ResNet and Transformer to extract the characteristic of building damage after disaster, and the text branch harnesses a BERT retriever for the text vectorization of posts as well as insurance policies and for the construction of a retrievable restoration index. To impose cross-modal semantic alignment, the model integrates a cross-modal interaction module to bridge the semantic representation between image and text via multi-head attention. Meanwhile, in the generation module, the introduced modal attention gating mechanism dynamically controls the role of visual evidence and text prior information during generation. The entire framework takes end-to-end training, and combines the comparison loss, the retrieval loss and the generation loss to form multi-task optimization objectives, and achieves image understanding and policy matching in collaborative learning. The results demonstrate superior performance in retrieval accuracy and classification index on damage severity, where the Top-1 retrieval accuracy has been improved by 9.6%.

CCS CONCEPTS

**Computing methodologies ~ Artificial intelligence ~ Computer vision ~ Computer vision representations** ~ Image representations

## 1 INTRODUCTION

Following natural disasters, such as earthquakes, hurricanes, and floods, homes of residents usually suffer different degrees of damage, and it is of great significance to assess these damages quickly and accurately. This process is not only relevant to the prompt settlement of insurance claims and the enforcement of policy conditions, but also holds considerable importance in helping the government and rescue department to assign funds resources and to take post disaster rehabilitation decisions [1]. Nevertheless, the traditional method of housing investigation and evaluation is mainly manually based on-site examination, which is not only cumbersome and time-consuming, but also likely to be influenced by the subjective judgments of the evaluator, and it has a common problem that the rating standards for different assessors are inconsistent and the results are subject to great deviation, seriously impeding the efficiency and accuracy of assessment, especially in the post-disaster large-scale situation.

---

[*] Corresponding author.

In recent years, along with the diversified ways of post-disaster image data acquisition (for example, the drone aerial photograph, the mobile terminal uploaded, etc.) and the digitization of the insurance documents storage, data-driven automatic damage evaluation tools have gradually emerged, providing a new way for improving efficiency and objectivity, and also promoting the change of the traditional experience depended to intelligent decision-oriented [2, 14].

In image perception area, the computer vision technology has been widely used in finding the damage of building70 structure. By analyzing multisource visual data (like satellite images, drone aerial images, and street view photos) with methods based on convolutional neural networks (CNNs) and visual Transformer models (viTs), researchers could classify with precision whether a house is complete and the extent of damage [3]. These measures have achieved good results in infrastructure recognition and damage level judgment, and hundreds of millions of RMB funds have been invested into the research.

The settlement of policy not only has to be attached with the information from the damage to the house, but also must have an insight into the contents of the terms of insurance contract terms, such as the scope of insurance liability, exemption clause, definition of the risk level and so on. Therefore, it is hard for one visual modality model to cover all the decision-making logic (i.e., ``supporting evidence with pictures and verifying responsibility with documents'') in the insurance scenarios, and it is of strategic importance to deeply integrate visual perception and text understanding to enable the formation of the cross-modal collaborative reasoning mechanism [4, 15-16].

To address the problem of the heterogeneous alignment and semantic complementarity between image-text modalities, the multimodal learning atmosphere has thrived in recent years and become a focused study area for representing visual or language information jointly. Particularly, in open-domain question answering and multimodal dialogue, the Retrieval-Augmented Generation (RAG) model performs well, as it integrates external knowledge retrieval and human speaking mechanisms, and can generate knowledge-grounded responses that are informative and context coherent [5]. Moreover, Liu [27] advocates for a broader Human-AI co-creation paradigm, emphasizing the role of collaborative design in intelligent systems—a perspective that aligns with our model's vision of integrating human-defined policies and AI-based damage perception. Unlike general vision-language pretraining (VLP) frameworks or existing multimodal RAG variants which primarily emphasize large-scale pretraining and general knowledge alignment. The performance improvements stem from three intertwined factors: (1) the hybrid ResNet-Transformer image encoder, which enhances local-global structural sensitivity; (2) the semantic retriever trained on real-world insurance policies, bridging document-level constraints; and (3) the end-to-end multi-task learning setup, jointly optimizing contrastive, retrieval, and generation losses[17-18].

## 2 RELATED WORK

Chengd et al. [6] presented a CNN-based architecture to Cascaded Convolutional Neural Network (CNN) for PDA by means of drone-captured post-hurricane images. They created a "DoriaNET" dataset gained during Hurricane Dorian, and implemented Earth Mover's Distance ($EMD^2$) loss rather than the conventional cross-entropy loss function, to more accurately express the ordinal magnitude which exists in damage levels. Romali and Yusop [7] developed systematic approach of flood damage and risk assessment of urban flood disaster (UFD) in Segamat, Malaysia. The annual expected damage (EAD) values for the residential and commercial districts are assessed based on flood depth and flood extent models (HEC-HMS and HEC-RAS plus ArcGIS), the exposure value of assets and the vulnerability curves.

Gourevitch et al. [8] note that when it comes to extreme weather and the housing market, the U.S. housing market "systematically underestimates" the risks of extreme weather events such as flooding. Via statistical analysis, they discovered that the market didn't fully account for future climate-related risks, like how higher flooding levels could push down house prices. Qin et al. [9] consider structural assessment and rehabilitation methods for high-rise concrete buildings under fire and explosion loads. An assessment method are given, the damage statistics, the structure of the damaged performance recovery material fire situation analysis, and puts forward a series of evaluation index and recovery strategy.

Günaydin et al. [10] observed that the majority of the damaged buildings were unreinforced masonry, utilizing either masonry or mud brick, were without horizontal and vertical reinforced concrete beams, and neither met the older nor new architectures based on seismic codes of Turkey, and collapsed according to typical failure modes at earthquakes such as diagonal cracks, wall inclination, and story dislocation of buildings. Rosenheimd et al. [11] developed an approach for incorporating household and housing unit attributes with critical infrastructure information for enhanced resilience modelling of communities.

Khajwal and Noshadravan [12] introduced a Post-Disaster Damages Assessment framework based on crowdsourcing. In order to address the problem of strong subjectivity and lack of reliability of crowdsourced data, the uncertainty and noise are handled in a maximum posterior estimation framework by statistically modeling both the uploaded public images and judgment opinions of damage. Wu et al. [13] learned pre-disaster and post-disaster image features through a double-path U-Net network, and integrated the attention modules into the skip connection, enabling the model to automatically pay attention to key spatial areas and enhance detection performance.

## 3 METHODOLOGIES

### 3.1 Image encoding is integrated with policy document retrieval

Input image $I \in \mathbb{R}^{H \times W \times 3}$ represents a color image of a post-disaster housing scene. First, we use the ResNet network to extract local features from the image, divide it into multiple visual patches, and map them to the vector space. In order to enhance the contextual modeling ability, the Transformer module is then introduced to further model the global dependencies between patches, as shown in Equations 1 and 2:

$$F_I^{(0)} = ResNet(I) \in \mathbb{R}^{N \times d}, \tag{1}$$

$$N = \frac{H}{p} \cdot \frac{W}{p}, \tag{2}$$

where $N$ is the number of patches the image is divided into, $d$ is the feature dimension, and $F_I^{(0)}$ is the initial image to represent the tensor. In order to capture the complex spatial pattern of structural destruction, we introduce Positional Encoding and superimpose it with image features, and send it into the Transformer structure to obtain the contextually enhanced image semantic embedding [19-20], as shown in Equation 3:

$$F_I = Transformer_{img}\left(F_I^{(0)} + PE\right) \in \mathbb{R}^{N \times d}, \tag{3}$$

where $PE$ is used to model the non-local relationship between different patches, and E is the final image semantic embedding sequence. The $PE$ represents the standard sinusoidal position encoding. $Transformer$ is used to model non-local relationships between different patches and $F_I$ represents the final image semantic embedding sequence. For each insurance policy document $P_j$, we use a pretrained BERT model to convert it into a fixed-length vector representation. Recent advancements by Zheng et al. have shown that comparative evaluation of multiple machine learning models can significantly improve representation quality when applied to structured document data [21]. These vectors will be used as semantic indexes to participate in the image-text matching retrieval process, given M policy documents, the encoding process of text branches is as follows in Equation 4:

$$v_j = BERT_{text}(P_j) \in \mathbb{R}^d, \quad j = 1,2,\dots,M, \qquad (4)$$

where $v_j$ represents the global semantic vector of the $j$-th policy, and the dimension is consistent with the image representation, which is used for cross-modal similarity calculation. To complete the image-text matching, we extract the query vector from the image branch, as Equation 5:

$$q = AvgPool(F_I) \in \mathbb{R}^d. \qquad (5)$$

Then, the similarity between the query vector and all policy semantic vectors is calculated, and the normalized cosine similarity is measured, as shown in Equation 6:

$$s_j = sim(q, v_j) = \frac{q^\top v_j}{\|q\|\|v_j\|}, \qquad (6)$$

where $s_j \in [-1,1]$ represents the similarity between the $j$-th policy and the image, which is used to construct the document candidate set. Select the top $k$ policy documents with the highest scores to form the search result set, as Equation 7:

$$\mathcal{D}_k = TopK\left(\{s_j\}_{j=1}^M, k\right), \qquad (7)$$

where, $\mathcal{D}_k$ will be used for the subsequent cross-modal fusion and generation process.

### 3.2 Retrieve enhanced generation and joint training mechanisms

The inputs to the generation module are image embedding and retrieved text embedding. We dynamically fuse the two modal information through the gating mechanism, and firstly, the semantic representations of the image and the text are averaged pooled, such as Equations 8 and 9:

$$z_I = AvgPool\left(\hat{F}_I^{(i)}\right), \qquad (8)$$

$$z_P = AvgPool\left(F_P^{(i)}\right). \qquad (9)$$

Then, construct the gating coefficient $\alpha$, which is used to dynamically balance the importance of the two modes, as shown in Equation 10:

$$\alpha = \sigma(W_g[z_I; z_P] + b_g), \quad \alpha \in [0,1], \qquad (10)$$

where $\sigma$ is the Sigmoid activation function, $[z_I; z_P]$ denotes the splicing operation. This mechanism is inspired by works such as Dai et al. and Feng et al., who demonstrate that attention gating in dialog systems and marketing models can effectively adapt modality-specific signals for optimal generation [22-23].

The final fusion modal is represented as a generator input, as shown in Equation 11:

$$z_{fused} = \alpha \cdot z_I + (1-\alpha) \cdot z_P, \qquad (11)$$

where $z_{fused}$ represents the decision-making semantics of the combination of images and texts. Use the Transformer decoder to combine the fusion representation $z_{fused}$ with the retrieved document set $\mathcal{D}_k$ as the context to generate a descriptive output, as in Equation 12:

$$\hat{y} = Decoder(z_{fused}, \mathcal{D}_k), \qquad (12)$$

The output sequence $\hat{y} = \{y_1, y_2, \ldots, y_T\}$ can be used as a description of the damage, a summary, or an assessment conclusion, depending on the task definition. Feng et al. demonstrated that reinforcement learning can enhance LLM-based output quality for specific tasks such as A/B testing, a technique we envision adapting for future damage feedback tuning [31]. The whole MM-RAG framework adopts end-to-end training to optimize the target integration of three types of loss functions. Wang et al. emphasize the importance of parallel optimization in large-scale recommendation systems, which aligns with our goal of scalable training efficiency [24], while Feng et al. propose latency-aware strategies for simultaneous learning of retrieval and generation, which inspire our loss design [25]. Contrastive loss is used to enhance the consistency of image and text representation, as follows Equation 13:

$$\mathcal{L}_{contrast} = -log \frac{\exp(\text{sim}(q, v^+))}{\sum_{j=1}^{M} \exp(\text{sim}(q, v_j))}. \tag{13}$$

Policy retrieval loss is used to evaluate the similarity ranking effect and is expressed as Equation 14:

$$\mathcal{L}_{retrieval} = \sum_{j \in \mathcal{D}_k} (s_j - y_j)^2, \tag{14}$$

where $y \in \{0,1\}$ indicates whether it is a matching policy. The generation loss is used to supervise text generation and is expressed as Equation 15:

$$\mathcal{L}_{gen} = -\sum_{t=1}^{T} log P(y_t | y_{<t}, z_{fused}, \mathcal{D}_k). \tag{15}$$

The combination of the three forms the final optimization objective function, as shown in Equation 16:

$$\mathcal{L}_{total} = \lambda_1 \mathcal{L}_{contrast} + \lambda_2 \mathcal{L}_{retrieval} + \lambda_3 \mathcal{L}_{gen}, \tag{16}$$

where $\lambda_1$, $\lambda_2$, $\lambda_3$ are weighting coefficients that are used to control the importance of different tasks. The "retrievable restoration index" is a hybrid document base, constructed from real-world insurance policy templates and disaster relief clauses.

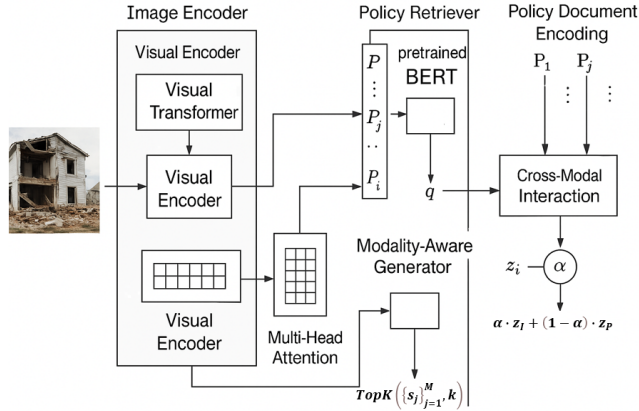

Figure 1. Architecture Diagram of MM-RAG Model.

The overall architecture flow of the Multimodal Retrieval-Enhanced Generation Framework (MM-RAG) proposed in this paper is depicted in Figure 1. The integration between ResNet and Transformer follows a serial fusion strategy. Specifically, ResNet-50 acts as the feature extractor that processes the input image into a grid of patch embeddings (e.g., 14×14 or 7×7), which are then flattened and fed into a Vision Transformer encoder. To preserve spatial positional relationships, we inject 2D sine-cosine positional encodings into the patch embeddings prior to

Transformer input. This ensures that both the relative layout of structural damage and global context are retained during self-attention operations, enabling robust representation of disaster-induced spatial cues.

## 4 EXPERIMENTS

### 4.1 Experimental setup

The experiment based on the multimodal data set xBD+Policy, the data was based on remote sensing disaster image set xBD, and was fused by the policy document constructed from the real insurance contract template, and the insurance document was used as an expert to establish the image text matching relationship. The dataset represents images across diverse natural disaster types (e.g., hurricanes, earthquakes and floods) and consists of pre-disaster images, post-disaster images, textual data of insurance policy and level-of-damage labels, motivating cross-modal representation learning from visual to textual. The stickers come in five levels and are designed in accordance with FEMA guidelines with text communicating coverage, exclusions and coverage amounts. We selected four representative comparison methods (Baselines) for performance comparison:

- The ResNet-50 + Softmax classifier (Visual-Only) takes post-disaster images and takes pre-trained ResNet-50 as feature extractor connected through fully connected layer for prediction of the damage level.
- BERT-Retriever + Rule-Based Matching (Text-Only) encodes each insurance policy document to a vector with pre-trained BERT, and infers the damage level indirectly through semantic matching (e.g., cosine similarity) or manually designed rules (such as the identification of insurance limit, exclusion clause).
- Late Fusion (ResNet + BERT) The multimodal joint use model follows a parallel strategy in the coding part of the image and text: the global visual feature of the image is extracted by the ResNet-50, and the text is encoded by BERT into semantic vectors, and then the two modal features are spliced and used as the input data of the fully connected layer for predicting the damage level.
- Text-Based Retrieval-Augmented Generation (Text-RAG) builds on the seminal RAG framework, taking natural language as input and mimicking the traditional retrieval-augmentation generation procedure. A question or description forms a query, a retriever retrieves a relevant document from a policy document library, and a generator generates a summary of the damage or an approximate grade according to the search.

### 4.2 Experimental analysis

In our experiments, the main evaluation metric is Accuracy which is calculated as the proportion of the count of samples predicted correctly to the total number of samples and is used to evaluate the overall performance of damage level classification.

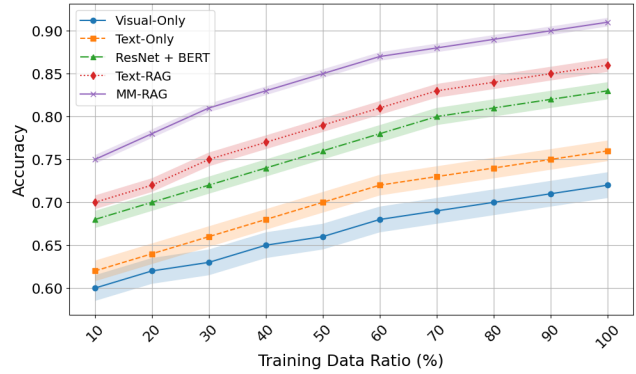

Figure 2. Accuracy Comparison with Error Margins across Methods.

The Figure 2 demonstrate that as the ratio of the amount of training data in the fine-tuning stage is increasing from 10%(or 20%) to 100%, the accuracy of all of the methods increases gradually, indicating that the more data, the better for learning. Among them, the unimodal models (Visual-Only, Text-Only) present only slight improvement, showing that it is hard to fully understand the meaning of damage with visual or textual condition alone. However, in medium data amount, ResNet+BERT and Text-RAG with simple fusion achieved better performance, so multimodal fusion and retrieval can contribute remarkable improvement. Our MM-RAG model is consistently ahead above 0.87 as early as 60% of the data, as well as tends to smooth convergence after 90%.

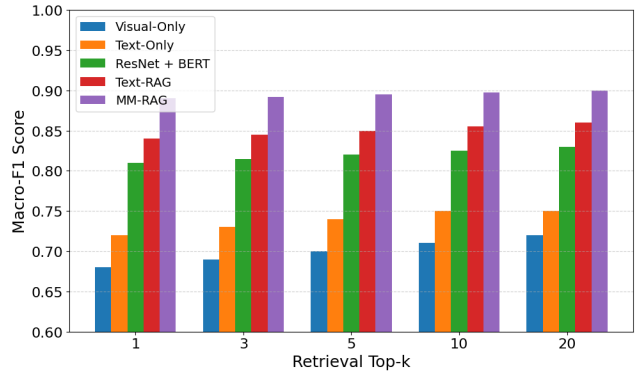

Figure 3. Macro-F1 Score vs. Retrieval Top-k across Methods.

Macro-F1 Score calculates the F1 value for each category, and then averages all the scores in arithmetic fashion to balance the differences of sample size between each damage level, which can more objectively reflect the global classification performance of the model for minority and majority class. From Figure 3, we can observe that with the increase of the retrieval width Top-k from 1 to 20, every method presents an increasing trend on F1 score, while the trend increase becomes more and more subtle, namely, more candidate documents can give more auxiliary information for document generation, but this improvement is gradually reduced.

Table 1. Top-5 Retrieval Accuracy with Embedding Dimension.

| Embedding Dimension | Visual-Only | Text-Only | ResNet + BERT | Text-RAG | MM-RAG |
|---|---|---|---|---|---|
| 64 | 0.55 | 0.65 | 0.72 | 0.75 | 0.82 |

| Embedding Dimension | Visual-Only | Text-Only | ResNet + BERT | Text-RAG | MM-RAG |
|---|---|---|---|---|---|
| 128 | 0.6 | 0.7 | 0.78 | 0.8 | 0.87 |
| 256 | 0.63 | 0.73 | 0.82 | 0.84 | 0.9 |
| 512 | 0.66 | 0.76 | 0.85 | 0.87 | 0.92 |

With the increase of the embedding dimension from 64 to 768, the Top-5 retrieval accuracy of each method has steadily improved but the increase has narrowed, and Table 1 shows that the high-dimensional vectors are helpful for semantic matching but there is a diminishing marginal return. MM-RAG has always been in the lead, jumping from 0.82 to 0.94, which is significantly better than Text-RAG, ResNet+BERT and unimodal methods.

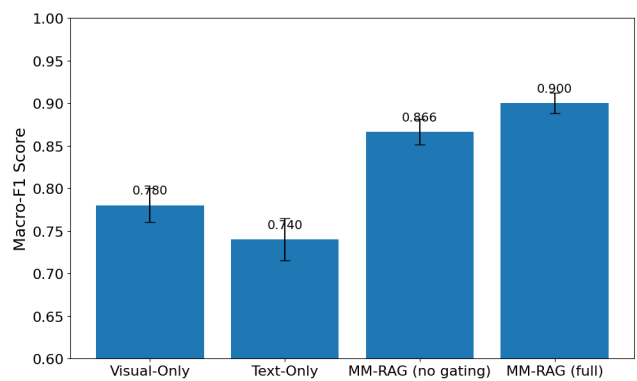

Figure 4. Ablation Simulation on Effect of Modal Attention Gating on Macro-F1.

Figure 4 shows that the unimodal baseline is limited, with Visual-Only and Text-Only Macro-F1 scores of only 0.78/0.74; removing the gating in MM-RAG increases the score to 0.866, highlighting the advantages of multimodal fusion; the complete MM-RAG, with the introduction of attention gating, reaches 0.900, with the lowest variance, achieving a gain of 3.8 percentage points compared to the non-gated version, validating the critical role of dynamically balancing visual and textual inputs to enhance the accuracy and stability of damage assessment.

## 5 CONCLUSION

In conclusion, the MM-RAG framework defined in this paper finally achieves deep fusion between images and policy documents for post-disaster housing damage assessment by joint application of ResNet-Transformer image coding, BERT text retrieval, cross-modal attention fusion and modal-aware gating generator and outperforms the single-modal and previous multimodal baselines on various testbeds. In the future, we can also introduce the time-series disaster evolution characteristic into this model, improve the semantic reasoning ability of complex policy clauses, and study online incremental learning and small-sample adaptation to expand the generalization. Recent work by Yan et al. [26] has shown that hybrid machine learning models can effectively capture temporal patterns in agricultural yield prediction, offering valuable insights for our integration of disaster evolution sequences.